\documentclass[sigconf, authorversion]{acmart}

\usepackage{color}
\usepackage{lipsum}
\usepackage{multirow}
\usepackage[normalem]{ulem}

\usepackage[T1]{fontenc}
\usepackage[english]{babel}
\usepackage{algorithm,algorithmicx,algpseudocode}
\usepackage{amssymb}

\usepackage{todonotes}
\usepackage{url}

\useunder{\uline}{\ul}{}

\acmConference[GECCO '18]{the Genetic and Evolutionary Computation
Conference 2018}{July 15--19, 2018}{Kyoto, Japan}
\acmYear{2018}
\copyrightyear{2018}
\acmArticle{4}
\acmPrice{15.00}

\begin{document}
\title[Data-efficient Neuroevolution with Kernel-Based Surrogate Models]{Data-efficient Neuroevolution\\ with Kernel-Based Surrogate Models}

\author{Adam Gaier}
\affiliation{%
  \institution{Bonn-Rhein-Sieg University of Applied Sciences}
  \institution{Inria / CNRS / Universit\'e de Lorraine}  
}
\email{adam.gaier@h-brs.de}

\author{Alexander Asteroth}
\affiliation{%
  \institution{Bonn-Rhein-Sieg University of Applied Sciences}
  \city{Sankt Augustin} 
  \country{Germany}
  \postcode{53757}
}
\email{alexander.asteroth@h-brs.de}

\author{Jean-Baptiste Mouret}
\affiliation{%
  \institution{Inria / CNRS /}
  \institution{Universit\`e de Lorraine}
  \city{Nancy} 
  \country{France}\postcode{54000}}
\email{jean-baptiste.mouret@inria.fr}

\begin{abstract}
Surrogate-assistance approaches have long been used in computationally expensive domains to improve the data-efficiency of optimization algorithms. 
Neuroevolution, however, has so far resisted the application of these techniques because it requires the surrogate model to make fitness predictions based on variable topologies, instead of a vector of parameters. 
Our main insight is that we can sidestep this problem by using kernel-based surrogate models, which require only the definition of a distance measure between individuals. 
Our second insight is that the well-established Neuroevolution of Augmenting Topologies (NEAT) algorithm provides a computationally efficient distance measure between dissimilar networks in the form of ``compatibility distance'', initially designed to maintain topological diversity. 
Combining these two ideas, we introduce a surrogate-assisted neuroevolution algorithm that combines NEAT and a surrogate model built using a compatibility distance kernel. 
We demonstrate the data-efficiency of this new algorithm on the low dimensional cart-pole swing-up problem, as well as the higher dimensional half-cheetah running task. In both tasks the surrogate-assisted variant achieves the same or better results with several times fewer function evaluations as the original NEAT.
\end{abstract}


\copyrightyear{2018} 
\acmYear{2018} 
\setcopyright{rightsretained}
\acmConference[GECCO '18]{Genetic and Evolutionary Computation Conference}{July 15--19, 2018}{Kyoto, Japan}
\acmBooktitle{GECCO '18: Genetic and Evolutionary Computation Conference, July 15--19, 2018, Kyoto, Japan}
\acmPrice{15.00}
\acmDOI{10.1145/3205455.3205510}
\acmISBN{978-1-4503-5618-3/18/07}

\maketitle


\section{Introduction}
\begin{figure}[ht]
	\centering
	\includegraphics{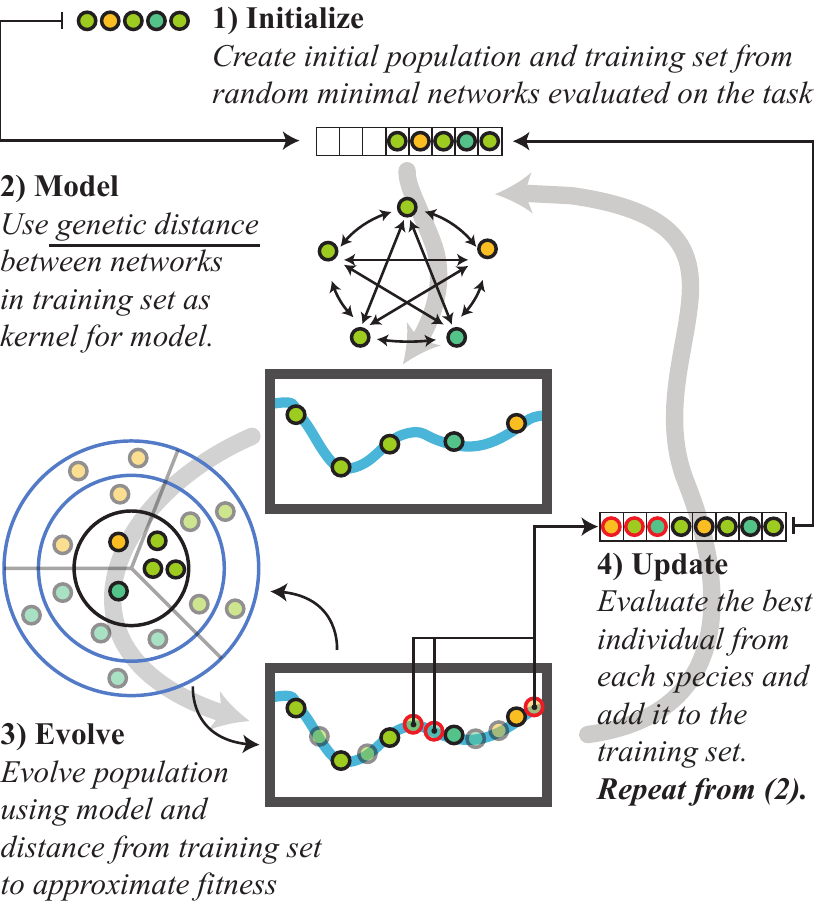}
	\caption
	{ 
		\textit{Surrogate-Assisted NEAT Overview}
		\newline
		Surrogate-Assisted NEAT integrates the compatibility distance, used by NEAT for speciation, into a kernel for Gaussian process regression. This compatibility distance kernel allows for fitness approximation of networks whose topologies vary and grow more complex. Integrating surrogate-assistance techniques into NEAT reduce the number of samples to reach the same performance by several times.
		}
	\label{fig:overview}
\end{figure}

Neuroevolution (NE), the optimization of neural networks through evolutionary algorithms, has proven to be effective in both machine learning~\cite{aaltonen2009observation, whiteson} and evolutionary robotics~\cite{neat, doncieux2015}, and its flexibility has made it a standard approach for experiments in embodied cognition~\cite{pfeifer2006body} and open-ended evolution~\cite{lehman2008exploiting,lehman2011abandoning}. 
Recent work has shown that even in deep neural networks, where millions of weights must be optimized, evolutionary techniques are a competitive alternative to gradient descent, demonstrating a surprising ability to scale~\cite{openAI_ne,uber_ESvSGD}. With a body of existing NE research, on topics such as exploration and overcoming deception, already being leveraged on deep learning problems~\cite{uber_GAandNovelty,uber_ESandNovelty}, NE is poised for a surge in interest.

The main challenge for deploying NE techniques in many applications is that they require many fitness evaluations --- too many for most realistic use cases. Deep neural networks, for instance, require long times to train even when large computational resources are brought to bear; in the case of robotics, there is a limit to the amount of interaction that is possible with the physical system.

A common approach to optimization in computationally expensive domains is to use approximate models of the objective function, or surrogate models~\cite{Jin2005,Forrester2009,Brochu2010,cully2015robots}. These models are created through an active learning process that aims at selecting points that are both promising in term of fitness and the likelihood to improve the predictions of the surrogate model. The typical loop alternates between selecting the best point to evaluate on the target system and retraining the model to take the new point into account. Machine learning techniques
are then used to construct surrogate models which map the genotype space to predicted fitness values~\cite{Forrester2009,Jin2005} .

Creating such a mapping is challenging when evolving neural networks, however: In cases where the topology and weights are both evolved, the dimensionality of the input space is not constant, and the dimensions themselves carry different meanings. Put differently, the surrogate model must be able to accept neural networks of varied layouts as an input, rather than a list of parameters.

Our first insight is that kernel-based methods, such as Gaussian process (GP) models and support vector machines, do not require that the inputs all have the same dimensions: they only require that some distance function between samples is defined. Distance measures designed for graphs, such as graph edit distance~\cite{ged} could theoretically be used to compute the distance between neural networks, but in practice are far too slow, with complexity exponential in the number of vertices. Though approximate measures of graph edit distance have been developed~\cite{neuhaus2006fast, riesen2009approximate}, even these are too slow for use as part of every prediction in an optimization algorithm.

Our second insight is that, when evolution is used to produce neural networks, we can glean additional information into the similarity of networks through their heredity. This is already done in the Neuroevolution of Augmenting Topologies (NEAT) algorithm~\cite{neat}, one of the most successful neuroevolution approaches. By tracking genes as they arise in the population, it is possible to create a meaningful and computationally efficient distance measure. NEAT uses this distance to cluster similar networks into species, \emph{here we propose its use as part of a kernel for Gaussian process regression}.

In summary, the primary idea explored in this work is that by tracing the common genes of networks as they evolve we gain a distance measure which can be used in a kernel-based surrogate model. 
Surrogate-assistance techniques can then be used to create a data-efficient neurevolution algorithm.

Broadly, the surrogate-assisted neuroevolution algorithm presented here proceeds as follows (Figure~\ref{fig:overview}, previous page):
	(1) a set of minimal networks are evaluated and form the initial training set and population,
	(2) the distance between all individuals in the training set is computed with a compatibility distance kernel, and a GP model constructed, 
	(3) the population is evolved with NEAT, with the fitness of individuals approximated by the compatibility distance model,
	(4) the best individuals in each species are evaluated and added to the training set, and the process repeats from (2).

\section{Related Work}
	\subsection{Neuroevolution of Augmenting Topologies}\label{sec:neat}	
	
Since its introduction in 2002~\cite{neat} NEAT has become the standard for neuroevolution. The core features of NEAT focus on overcoming the competing conventions problem of dissimilar networks. The algorithm begins with a population of minimal networks, which grow more complex through mutation. Whenever new nodes and connections are added to the network they are given unique markers. These markers allow common components of dissimilar networks to be identified, providing a basis for crossover and the clustering of networks into species. Species compete against each other for a share of the total offspring they contribute to the next population, and individuals compete within species to provide those offspring.

NEAT has seen successes in domains from video game AI~\cite{stanley2005evolving} to particle physics~\cite{aaltonen2009observation}, and forms the basis and inspiration for a host of other innovations. It is the underlying algorithm for the evolution of compositional pattern producing networks~\cite{stanley2007compositional} which were in turn applied to the indirect encoding of large scale networks with the HyperNEAT algorithm~\cite{Stanley2009}. The ability of NEAT to produce networks of increasing complexity has also made it an ideal tool for exploring open-ended evolution and novelty-based search~\cite{lehman2008exploiting, lehman2011abandoning}.


	\subsection{Gaussian Process Models}\label{sec:gp}

\begin{figure*}[ht]        
    \centering        
    \includegraphics[width=1\textwidth]{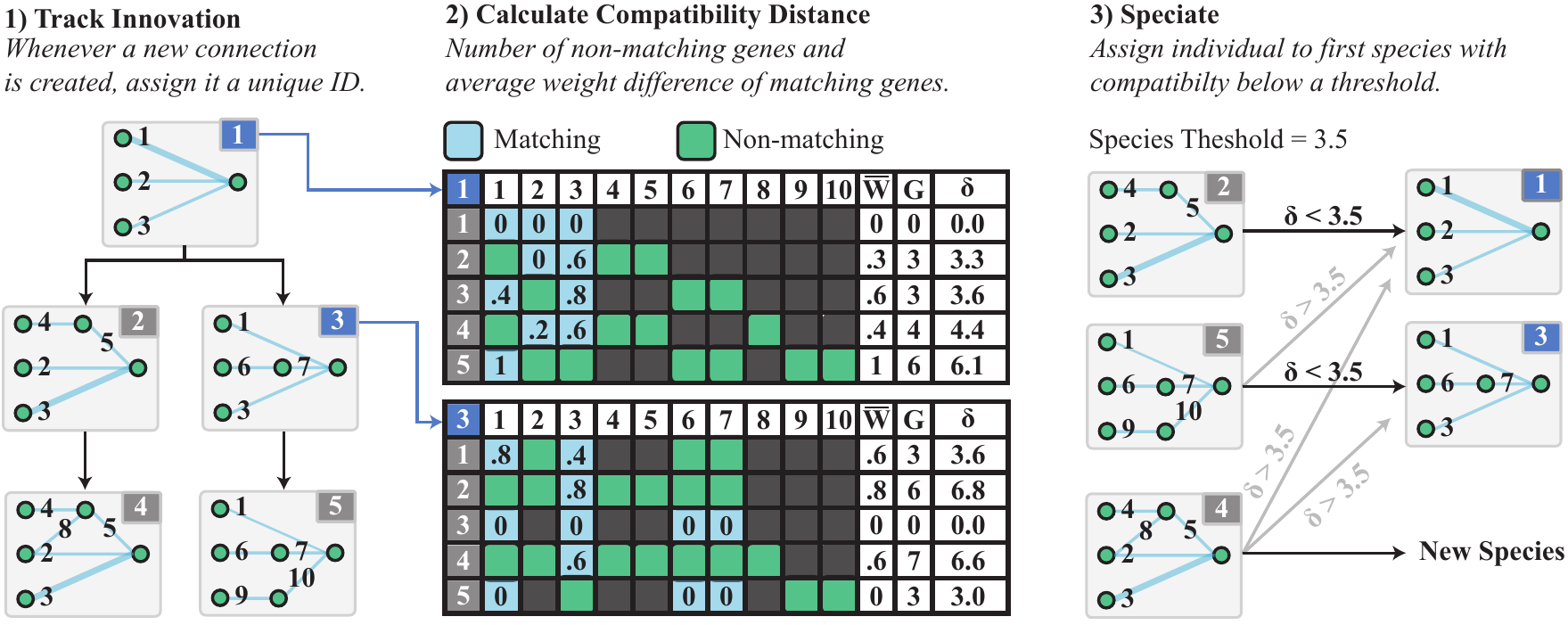}       
    \caption      
    {     
        \textit{Calculating Compatibility Distance in NEAT}       
         \newline      
         By tracking genes as they arise NEAT allows for the efficient comparison of dissimilar networks. The ``compatibility distance'' between two networks is a weighted sum of the number of genes they do not share and the weight differences between the genes they do. Each species has a representative. Individuals are compared to these representatives, and are assigned to the first species whose representatives compatibility distance is below a threshold. If none exist the individual forms a new species.
    }         
    \label{fig:neat}      
\end{figure*}

Surrogate models can be constructed with a variety of machine learning techniques~\cite{Jin2005,Forrester2009}, but GP models~\cite{Rasmussen2006} are most commonly used in modern approaches. GP models are accurate even with small data sets, and include a measure of uncertainty in their predictions, important for balancing exploration and exploitation.

Gaussian process models use a generalization of the Gaussian distribution: where a Gaussian distribution describes random variables, defined by mean and variance, a Gaussian process describes a random distribution of functions, defined by a mean function $\mu$, and covariance function $k$. 
\begin{equation}
f(x) \sim GP(\mu(x), k(x,x'))
\end{equation}
GP models are based on assumptions of smoothness and locality: the intuition that similar individuals will have similar behavior. A covariance function $k$ defines this relationship precisely in the form of a kernel. A common choice of kernel is the squared exponential function: as points $x$ become closer in input space they become exponentially correlated in output space:
\begin{equation} \label{eq:sqExp}
    k(\mathbf{x_i,x_j}) = \exp{ \Big( -\frac{1}{2} \|\mathbf{x}_i - \mathbf{x}_j\|^2 \Big)}
\end{equation}
Given a set of observations $D = ({x_{1:t}, f_{1:t}})$ where $f_{1:t} = f(x_{1:t})$,  we can build a matrix of covariances. In the simple noise-free case we can then construct the kernel matrix:
\begin{equation}
    K = 
     \begin{bmatrix}
        k(x_1,x_1)    & \cdots    & k(x_1,x_t)    \\
        \vdots        & \ddots    & \vdots        \\
        k(x_t,x_1)    & \cdots    &  k(x_t,x_t)
    \end{bmatrix}
\end{equation}
When considering a new point ($x_{t+1}$) we can derive the value ($f_{t+1} = f(x_{t+1})$) from the normal distribution:
\begin{equation}
    P(f_{t+1}|D_{1:t},x_{t+1}) = 
    \mathcal{N} \Big( \mu_t (x_{t+1}), \sigma_t^2 (x_{t+1}) \Big)
\end{equation}
            where:               
\begin{align}
    \mu_t(x_{t+1}) &= \mathbf{k}^T \mathbf{K}^{-1} \mathbf{f}_{1:t} \\
    \sigma_t^2(x_{t+1}) &= k(\mathbf{x}_{t+1}, \mathbf{x}_{t+1}) - \mathbf{k}^T \mathbf{K}^{-1}\mathbf{k}
\end{align}
gives us the predicted mean and variance for a normal distribution at the new point $x_{t+1}$. When the objective function is evaluated at this point, we add it to our set of observations $D$, reducing the variance at $x_{t+1}$ and at other points near $x_{t+1}$.

		\subsubsection*{Bayesian Optimization} \label{sec:bo}

Modern surrogate-assisted optimization often takes place within the framework of Bayesian optimization (BO)~\citep{Brochu2010, Calandra2013, cully2015robots, Shahriari2016, pautrat2017, gaier2017data}. BO approaches the problem of optimization not only as one of finding the most optimal solution, but of modeling the underlying objective function in high performing regions. 

BO requires a probabilistic model of the objective function, and so GP models are typically employed. This model is used to define an acquisition function, which describes the utility of sampling a given point. The objective function is evaluated at the point with maximal utility and added to the set of observations. The updated observation set is used to rebuild the model, and the process repeats.


In this work, we use the \textit{upper confidence bound} (UCB) acquisition function~\citep{Srinivas2009}. A high mean~($\mu$) and large uncertainty ($\sigma$) are both favored, with relative emphasis tuned by the parameter $\kappa$:
\begin{equation} \label{eq:UCB}
  UCB(\mathbf{x}) = \mu(\mathbf{x}) + \kappa\sigma(\mathbf{x})
\end{equation}

UCB performs competitively with more complex acquisition functions such as Expected Improvement (EI) and Probability of Improvement (PI) \citep{Brochu2010,Calandra2013}. 



\begin{figure*}[ht]
    \centering
    \includegraphics[width=1\textwidth]{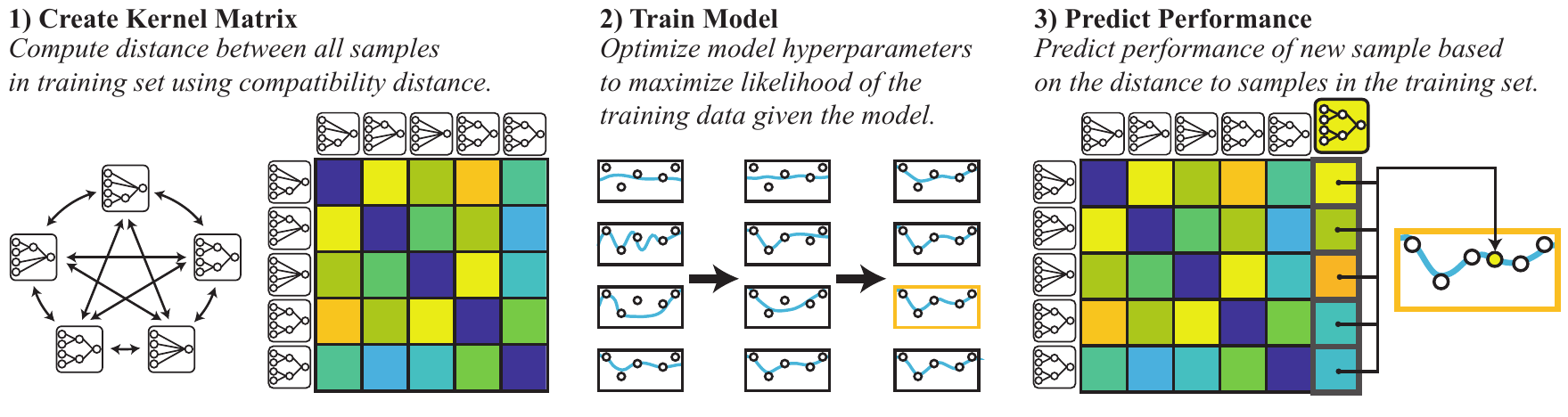}
    \caption
    {
        \textit{Predicting Performance from Compatibility Distance}
        \newline
        To predict performance based on compatibility distance a kernel matrix must first be created by comparing every individual in our training set to every other with the kernel function. We then train the model hyperparameters to maximize the likelihood of the training set given the model. Fitness for an unknown individual can then be predicted using the compatibility distance from each individual in the training set and the found model hyperparameters.
    }
    \label{fig:cd-gp}
\end{figure*}

\section{Compatibility Distance Kernel GP}\label{sec:cd-gp}

We would like to use GP models to approximate the fitness function for a surrogate-assisted neuroevolution algorithm.
Though the estimates of GPs are typically based on distance between samples in parameter space, they are a kernel-based method, and as such only require some distance measure between samples. 


In the case of neuroevolution, where the topology of the network is evolved along with the values of the weights, we do not have a static or consistent parameter space. As the population of networks grow and change, the genotype spaces they exist in diverge into varied dimensions with inconsistent meanings, leaving standard distance measures such as Euclidean distance unusable.

If a meaningful distance measure was found, then GP models could be used. Neural networks are a class of directed graph, and there already exist measures to compare graphs, such as graph edit distance~\cite{ged}. Unfortunately, even approximate graph edit distances are too expensive to compute for every prediction~\cite{neuhaus2006fast, riesen2009approximate}. 

As we are producing the networks through an evolutionary process, however, we can track the phylogenetic links between individuals and compute a distance between them based on their common genes.
NEAT introduces just such a mechanism by assigning \emph{innovation markers} whenever a new gene arises. The genome of a neural network evolved with NEAT is composed of a list of nodes and a list of connections. Starting with a fully connected minimal network new nodes are added by splitting existing connections, adding a new node which has a connection from the source node and to the destination node. New connections can then be added to and from this node through mutation. In either case, whenever a connection is added it is assigned a unique \emph{innovation number}, implemented simply as a running counter (Figure~\ref{fig:neat}, left).

By comparing these identifiers similar structures in dissimilar genotypes can be easily and efficiently identified, allowing the distance between two individuals to be calculated (Figure~\ref{fig:neat}, center). This \emph{compatibility distance} is used by NEAT to cluster similar individuals into species, and we can use it as a kernel for our GP, allowing us to perform predictions across dissimilar structures.

The canonical NEAT \cite{neat} introduces several coefficients and normalization factors which provide additional degrees of freedom in how exactly this value is calculated, but we simplify it here to:
\begin{equation}
\delta(\mathbf{x_i, x_j}) = c_1 \cdot G(\mathbf{x_i, x_j}) + c_2 \cdot \overline{W}(\mathbf{x_i, x_j})
\end{equation}
where the compatibility distance $\delta$ between two individuals $x_i$ and $x_j$ is the weighted sum of the number of non-matching genes $G$ and the average weight differences of matching genes $W$. The compatibility distance is used by NEAT to cluster individuals into species. New individuals are compared to representatives of each species found in the previous generation, and join the first whose distance is below a certain threshold (Figure~\ref{fig:neat}, right). 

To produce the kernel matrix of the GP we use a compatibility distance kernel function which returns the squared exponential compatibility distance between samples:
\begin{equation} \label{eq:compK}
    k(\mathbf{x_i,x_j}) = \exp{ \Big( -\frac{1}{2} \delta(\mathbf{x}_i, \mathbf{x}_j)^2 \Big)}
\end{equation}
While precision of the predictions when only matching connections and weight differences may be limited, the underlying assumption, that the more similar two individuals are the more similarly they can be expected to behave, holds. The rough predictions produced by the predictive model provide enough information to ensure that higher fitness individuals produce more offspring.

To train a GP model, its hyperparameters are tuned to make the known observations most likely given the model, balancing accuracy and simplicity. We tune two hyperparameters of our kernel: the characteristic length scale ($\ell$), which can be thought of as the distance you can move in input space before the output value changes significantly, and the variance ($\eta$), how far the output signal varies from the function's mean. Integrating these hyperparameters give us a kernel of the form:
\begin{equation} \label{eq:compK2}
    k(\mathbf{x_i,x_j}) = \eta\exp{ \Big( -\frac{1}{2\ell} \delta(\mathbf{x}_i, \mathbf{x}_j)^2 \Big)}
\end{equation}
These hyperparameters $\theta$ are optimized by maximizing the log likelihood of the fitness values $\mathbf{y}$ given the individuals in the population $\mathbf{x}$ and compatibility kernel matrix $\boldsymbol{K}$:
\begin{equation}
    \text{log }p(\boldsymbol{y}|\boldsymbol{x},\boldsymbol{\theta}) = 
    -{1 \over 2}\boldsymbol{y}^T(\boldsymbol{K-\sigma_n\boldsymbol{I}})^{-1}\boldsymbol{y}
    -{1 \over 2}\text{log}~|(\boldsymbol{K-\sigma_n\boldsymbol{I}})|
    -{n \over 2}\text{log}2\pi
\end{equation}

Typically, gradient-based optimization is used to maximize the likelihood, but this is not possible here because the compatibility distance is not differentiable. Instead, we use the Covariance Matrix Adaptation Evolution Strategy (CMA-ES), which has been proven as effective at optimizing Gaussian process model parameters as gradient-based methods in other contexts~\cite{blackdrops2017}. In addition to the kernel specific parameters $\ell$ and $\eta$ the mean ($\mu$) and signal noise ($\sigma$) are also tuned.

The training and prediction process can be summarized as follows: all individuals in the training set are compared using NEAT's compatibility distance metric to produce a covariance matrix of their similarity, the hyperparameters of the model are then optimized with CMA-ES to maximize the likelihood of the data given the model, and finally a prediction can be calculated based on the model and distance to the individuals in the training set. This training and prediction process is illustrated in Figure \ref{fig:cd-gp}.

\section{Surrogate-Assisted NEAT}

\begin{figure*}[ht]
  \centering \includegraphics[width=1\textwidth]{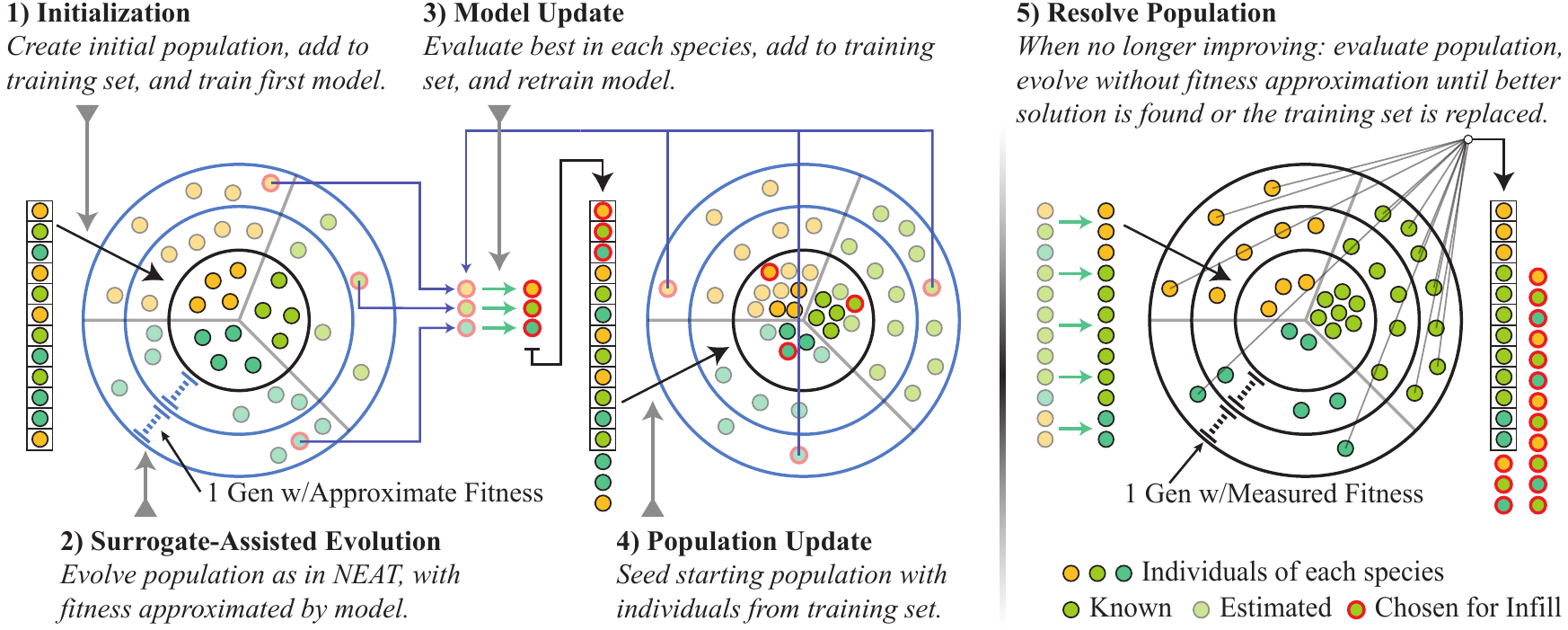}
    \caption
    {
        \textit{Surrogate-Assisted NEAT}
        \newline
        1) An initial population of minimal networks is used as a training set, serving as the basis for the initial surrogate model. 
        \newline
        2) The population evolves according to the NEAT algorithm, with fitness replaced by the approximations of the model. 
        \newline
        3) From each species, the individual with the highest predicted performance is selected and evaluated on the task. These new samples are placed in the training set, replacing the oldest samples if the training set has reached its maximum size.
        \newline
        4) The training set and population are combined to form a new population, and the process repeats from step 2.
        \newline
        5) When fitness is no longer improving, the entire population is evaluated and added to the training set. If none of these individuals are an improvement, a new generation is produced with NEAT, which is evaluated and added to the training set. Evolution continues in this way until a) a better individual is found, or b) the entire training set has been replaced with new individuals. The model is then retrained and the algorithm resumes from (2) with this new population and training set.
    }
    \label{fig:saNeat}
\end{figure*}

Predictions based on a GP model with a compatibility distance kernel can identify the most promising individuals to test and the most promising genotype regions to explore. By judiciously sampling these individuals we can improve the accuracy of our models in optimal regions and perform the same simultaneous topology search and weight optimization as NEAT, with a focus on data-efficiency. The core algorithmic machinery of NEAT is maintained, with the adjustments needed to place NEAT into a surrogate-assisted framework outlined below and illustrated in Figure~\ref{fig:saNeat}.

\paragraph{Initialization}
This surrogate-assisted variant of NEAT begins just as the original version of NEAT, by initializing a set of minimal networks and testing them. These initial samples and their fitness form the training set of our first model. The distance between all samples is computed and a Gaussian process model trained.

\paragraph{Surrogate-Assisted Evolution}
The population is evolved according to the standard mechanisms of NEAT: individuals are grouped into species, a number of offspring are assigned to each species based on their fitness, and finally tournament selection and variation is performed within each species to produce a new population. The compatibility distance between the newly produced individuals and all individuals in the training set is then calculated. Based on the model and this distance, we calculate the utility of sampling each new individual. We reward individuals with high predicted fitness and high uncertainty, using the UCB acquisition function (see Section~\ref{sec:bo}). We then repeat the evolutionary process, grouping the new individuals into species, and using this utility value in place of fitness when assigning offspring to species and determining the winners of intra-species tournaments.

\paragraph{Model Update} 
Surrogate-assisted evolution is repeated a number of times before new samples are added to the model. When selecting these new samples we take advantage of NEAT's concept of species. The species clustering in NEAT ensures that a diversity is maintained, and new complexity nurtured. Species are clustered using the same measure of similarity as our model, and so by sampling one individual in a species we improve the prediction accuracy on other individuals in the same species. To improve our model across species the best performing individual in each is evaluated on the task, added to the training set, and the model retrained.

The training set is limited to a maximum size, and if adding new samples would extend it beyond that size the oldest samples are replaced. This sliding window approach to our training data serves dual purposes. The first is to keep our models relevant to the current individuals being evolved. As the genotypes become more complex the distance from older, simpler individuals becomes less relevant. Older individuals will not only have lower fitnesses, but as the population explores new spaces older individuals will contain many genes which do not exist in the current population, providing little benefit to prediction.

There is also a computational advantage in a smaller training set. New individuals must be compared to every individual in the training set, and the matrix of distances between the training set samples must be inverted when creating the Gaussian process model, an operation with complexity cubed in the number of samples~\cite{Rasmussen2006}. A limited training set of recent individuals ensures a computationally efficient model focused on relevant and high performing regions.

	\begin{figure*}[ht]
	\centering
	\includegraphics[width=1\textwidth]{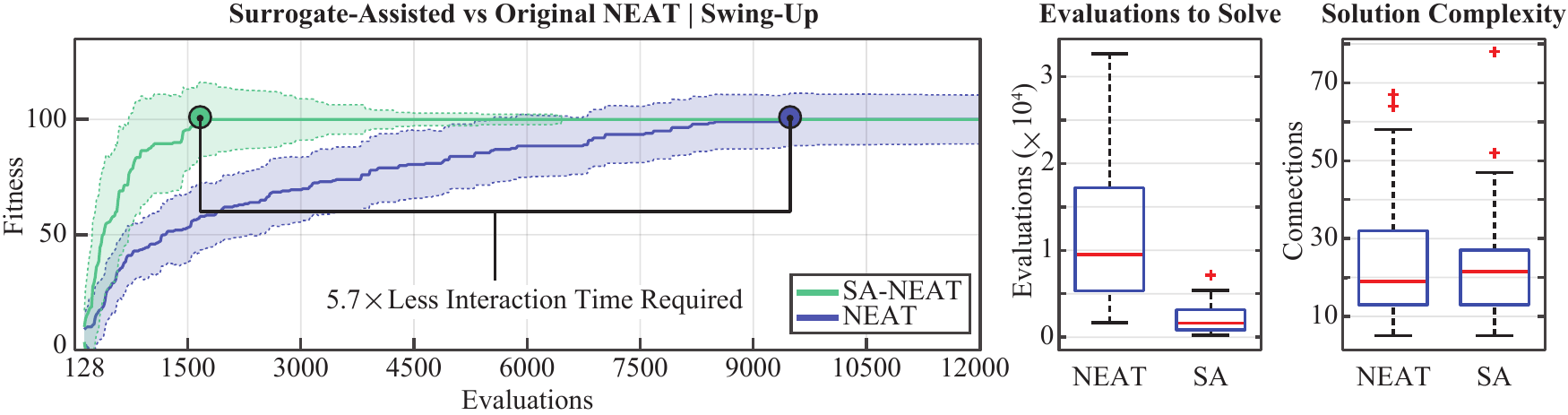}
	\caption
	{
		\textit{Cart-Pole Swing-Up: Evaluations to Solve}\newline
		Comparison of median best fitness found over 50 trials of SA-NEAT and NEAT on the cart-pole swing-up task. Shaded region indicates one standard deviation from the median. SA-NEAT solved the task in a median of 1664 evaluations, while NEAT required 9488, nearly six times the number of evaluations. Even the most data-efficient quartile of of NEAT runs required as many evaluations as the least data-efficient SA-NEAT runs. The complexity of solutions found in each case are similar, demonstrating that the same variable dimensional space explored by NEAT is also explored in the surrogate-assisted variant.
	}
	\label{fig:swingPlot}
\end{figure*}

\paragraph{Population Update}
The training set serves another purpose, doubling as a store of known starting points for evolution. As generations of surrogate-assisted evolution repeat, the population drifts farther away from known solutions where reasonable predictions can be made. In typical cases of surrogate-assisted optimization this is not a concern: all solutions occupy the same space and predictions become more accurate as the solution space is explored. With a complexifying genome, however, new dimensions are introduced faster than they can be efficiently explored.

To contain this explosion of genotypic complexity we reintroduce known samples back into the population. Whenever we update the model, we also add one member of the training set for every member of the population, with the most recent added first, effectively doubling the breeding pool for that generation. This larger collection of individuals is divided into species and recombined to form a new population of the standard size. Much of the new population will have known samples as one or both parents, pulling the population back towards known genetic dimensions, allowing more accurate predictions of their fitness.

\paragraph{Resolve Population}
In cases where the parameter space is fixed, surrogate-assisted methods will reliably converge on the optima as more samples are obtained, but in an open-ended space this is not the case. In the event that the algorithm converges on a local optimum and stagnates we ``resolve'' the population, replacing fitness approximations with true fitness values.

If enough newly added individuals are added to the training set without improvement, the entire population is evaluated on the task, revealing any individuals in the population which would achieve higher fitness but were never chosen for evaluation. 
If no better solutions are found then the speciation and recombination of NEAT repeats, and the entire population again is evaluated on the task. Every individual evaluated is added to the training set, and this NEAT evolution continues until either a better solution is found or the entire training set is replaced with new individuals. At that point the GP model is reconstructed and the algorithm begins again with a diverse, complex, but known population. With the search space once again well-modeled the process of surrogate-assisted evolution, model update, and population update resumes.

\section{Experimental Results}
	\subsection{Cart-Pole Swing-Up}
		\begin{figure*}[ht]
	\centering
	\includegraphics[width=1.0\textwidth]{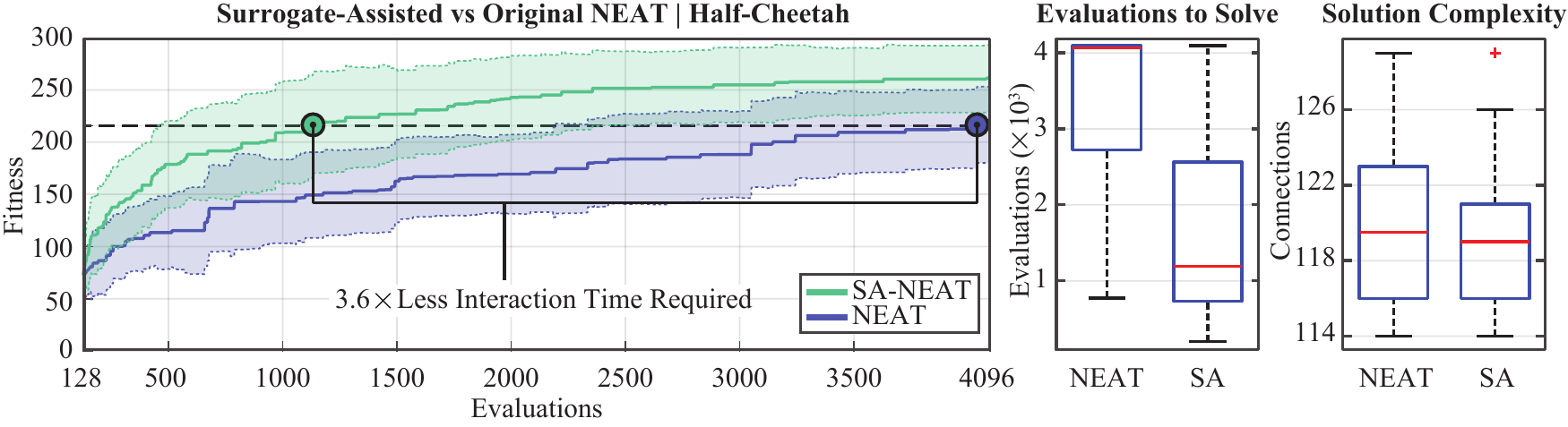}
	\caption
	{
		\textit{Half Cheetah: Best Fitness Per Evaluation}\newline
		Comparison of median best fitness found over 30 trials of SA-NEAT and NEAT on the half-cheetah running task. Shaded region indicates one standard deviation from the median. With no solve state, data-efficiency was compared by the evaluations required to reach the highest median fitness reached by NEAT. SA-NEAT reached the median level of fitness achieved by NEAT in 4096 evaluations in a median of 1136 evaluations, a difference of more than three and a half times. Even the most data-efficient quartile of NEAT runs required as many evaluations as the least data-efficient SA-NEAT runs. Again, the complexity of the networks produced by the two approaches was equivalent.
	}
	\label{fig:halfCheetah}
\end{figure*}

	


\paragraph{Setup and Hyperparameters}
We test our approach to surrogate-assisted neuroevolution first on a classic benchmark control problem, the cart-pole swing-up. The system begins with a cart on a two dimensional track with a pole hanging below it, with the objective of swinging the pole into an upright position and maintaining it in a balanced state. This task is more difficult than benchmarks used in many evolutionary computation publications, such as pole-balancing, and cannot be solved with a linear controller~\cite{raiko2009variational}, requiring networks to grow beyond their initial minimal state. 

The known state of the system is the position and velocity of the cart, and the angle and angular velocity of the pole. Inclusion of a bias node results in a total of 5 inputs, with a single output node specifying a command to the system as a percentage of the maximum force. The cart-pole system used here is composed of a 2 kg cart and a 0.5 m pole weighing 0.5 kg. The maximum force which can be applied to the cart is 10 N, with control signals sent to the system at every 0.25 seconds, for a total of 5 seconds.

Controllers are rewarded for the most consecutive time steps in which the pole is held upright. If, for example, the pole is held upright for 25 time steps, falls, then is swung back up and held upright for an additional 15 time steps the controller is only awarded a fitness of 25, not 40. Fitness is only awarded for time steps in the second half of the trial, for a maximum fitness of 100.
	 
NEAT has a large number of hyperparameters, too many to test and tune exhaustively. Instead we conducted preliminary testing with different levels of variation per generation, based on the hyperparameters presented in the canonical NEAT article~\cite{neat}. The probabilities to add nodes and connections, reenable nodes, perform crossover, and mutate weights were scaled by $2, 1, {1 \over 2}, {1 \over 4},$ and ${1 \over 8}$: preliminary tests showed NEAT's best performance when variation was scaled by ${1 \over 2}$ and so these hyperparameter values were used. 
	
In runs of SA-NEAT, 4 generations of evolution took place before selecting 4 infill individuals to add to the population. These were taken from the top 4 species, except in the case where less then 4 species were present, in which case the highest performing individuals were taken in their place. A training set of 512 individuals was maintained, and the population ``resolved'' if 128 individuals were added to the training set without improvement. To keep the same amount of variation in one sampling iteration of SA-NEAT as would occur in a single generation of NEAT, the rates of variation were decreased by ${1 \over 4}$  (${1 \over 8}$ of the hyperparameters in~\cite{neat}). Table \ref{table:params} outlines the hyperparameters and their relationship.


\begin{table}[]
\centering
\caption{SA-NEAT Hyperparameter Values and Derivation}
\label{table:params}
\begin{small}
\begin{tabular}{lll}
\multicolumn{1}{c}{{\ul Hyperparameter}} & {\ul Relative Value}       & {\ul Absolute Value} \\
\# of Species                            & -                          & 4                    \\
Gens Per Infill                          & -                          & 4                    \\
Population Size                          & -                          & 128                  \\
Variation                                & Base  / Gens Per Infill    & (Published / 8)      \\
Inds Per Infill                          & \# of Species              & 4                    \\
Training Set                             & Inds Per Infill $\times$ Pop Size & 512                  \\
Stagnation                               & Population Size            & 128                 
\end{tabular}
\end{small}
\end{table}

\paragraph{Results}
	The comparison of performance between NEAT and SA-NEAT on the swing-up task over 50 replicates is shown in Figure~\ref{fig:swingPlot}. We compare only the number of fitness evaluations performed on the cart-pole simulator: fitness predictions using the surrogate model are not counted. A dramatic speed-up can be observed: by the time NEAT exhausted evaluations equal to 13 generations, half of SA-NEAT runs had already solved the task. This represents a gain in data-efficiency of nearly six times. The acceleration is made even more stark when the full distribution of results is examined. Even the most data-efficient quartile of NEAT replicates require as many evaluations as the least data-efficient runs of SA-NEAT. It should also be noted that the complexity of the produced networks matches those found by NEAT, illustrating that SA-NEAT is indeed exploring the same variable dimensional space as NEAT.


	\subsection{Half-Cheetah}

	
	


\paragraph{Setup and Hyperparameters}
To test the SA-NEAT approach on a higher dimensional problem, we compare its performance in the half-cheetah running task. The half-cheetah system described in~\cite{wawrzynski2007learning} is one half of a quadruped robot with a front and back leg, with each leg consisting of 3 joints. The system has a state space of 17 values: the velocity and angles of each joint, the position and velocity of the body in the $x$ (forward) and $z$ (up) directions, and the angle and angular velocity of the body in $y$ (side). Each of the 6 joints are controlled by sending a torque command, for a total 108
weights in the initial minimal networks (including a bias input).

To encourage smoother gaits we prevent rapid direction shifts in joint direction by applying the output of the neural network not as raw joint torques, but as adjustments to the existing torque levels. Torques on each joint at a time step $t$ is applied as:
\begin{equation} \label{eq:torque}
    \mathbf{torque}_{(t)} = \mathbf{torque}_{(t-1)} + \mathbf{y}
\end{equation} 
where $\mathbf{y}$ is the output vector of the neural network.

The OpenAI gym implementation~\cite{aiGym} of the half-cheetah is run for 150 time steps\footnote{This is significantly fewer time steps than is typical for this task in reinforcement learning experiments, and so results are not directly comparable to those in the literature. The purpose of these experiments is only to establish the benefits w.r.t. NEAT: more thorough comparisons with previous work will be presented in a future publication.}
, with fitness awarded for moving forward with minimal effort:
\begin{equation} \label{eq:fitness}
	         \text{fitness} = \sum\limits_{t=1}^{150} (\text{pos}_{t+1}-\text{pos}_t)/dt - 0.1 \times \mathbf{torque}_{t}^2	          
\end{equation}

The same hyperparameters used for NEAT and SA-NEAT in the swing-up task were used here. As there is no half-cheetah solve state, and is much more expensive to simulate than the cart-pole, we limit the number of evaluations to 4096 and run only 30 replicates.

\paragraph{Results}
Even in this more complex problem, SA-NEAT outperforms NEAT (Figure~\ref{fig:halfCheetah}), reaching the same levels of fitness as NEAT at the end of the trial using only a third of the needed evaluations. This not only confirms our earlier experiment, but also shows that SA-NEAT is able to navigate a high dimensional weight space as well as searching the space of possible topologies.

While the swing-up benchmark is not a trivial task the space of solutions, even in the complexifying case, is relatively limited. With a minimal topology of five inputs and one output it begins as only a five dimensional problem. The half-cheetah, on the other hand, begins in a space that is more than 100 degrees of freedom. As the compatibility distance kernel is independent of the dimensionality of the underlying genotype, our models are still able to make useful predictions in this space with only a few hundred samples.

\section{Conclusion and Discussion}
We introduced a surrogate-assisted variant of NEAT, SA-NEAT, as a data-efficient method of performing neuroevolution in computationally expensive problems. By taking advantage of the phylogenetic information produced as a byproduct of the evolutionary process, we created a new kernel to judge similarity of neural networks based on their shared genes. Using GP models built with this kernel we are able approximate the performance of individuals, allowing us to achieve similar results with several times fewer evaluations. Fewer evaluations does not guarantee a faster experiment: when the fitness function is evaluated in simulation, there is always a trade-off between the cost of modeling and evaluation. By limiting the model to recently evaluated samples we boundnd its complexity, and it is possible that both accuracy and performance could be further improved with even more purposefully constructed models.



In the presented approach, though species diverged into varied and distant genealogies, a single training set and model were used for prediction. Due to the squared exponential relationship in the kernel, individuals in more distant species have little if any effect on the predicted performance, yet are still considered in the comparison. Producing surrogate models with individuals only from within a single species would reduce the needed number of comparisons. Apart from computational concerns, species specific models could also have more predictive power, as the hyperparameters of the model could more accurately reflect the particular genotype region, rather than their likelihood over the entire training set.

NEAT is used to evolve compositional pattern producing networks (CPPNs)~\cite{stanley2007compositional} , indirect encodings used to produce neural networks~\cite{Stanley2009}, images~\cite{picbreeder, nguyen2015deep}, and solid objects~\cite{clune2011evolving}. 
Whether our approach is as successful in evolving indirect encodings remains to be seen, but as many engineering domains rely on expensive simulations, a data-efficient method of evolving CPPNs would allow their application in real world design problems.


As neuroevolution gains increased attention from industry for its capabilities in large scale problems, 
the tasks it is charged with will only grow in complexity.
Despite the continued growth in computing power there will always be demand for more, and this is especially true for population-based approaches.
Combining data-efficient machine learning with neuroevolution ensures that the 
diversity preserving,
novelty seeking, 
deception avoiding
abilities of evolutionary approaches can still be applied, regardless of the complexity of the challenges presented.

\section*{Acknowledgments}
This work received funding from the European Research Council under the European Union's Horizon 2020 research and innovation programme (grant agreement number 637972, project ``ResiB bots'') and the German Federal Ministry of Education and Research under the Forschung an Fachhochschulen mit Unternehmen programme  (grant agreement number 03FH012PX5 project ``Aeromat'').

\bibliographystyle{ACM-Reference-Format}
\bibliography{saNeat} 

\end{document}